\documentclass[12pt]{article}

\usepackage{graphicx}
\usepackage{amsmath,amssymb}
\usepackage{orcidlink}
\usepackage{booktabs} 
\usepackage{lineno} 
\usepackage{float} 
\usepackage{authblk}
\usepackage{dsfont} 

\begin{document}

\title{Event-Driven Neuromorphic Vision Enables Energy-Efficient Visual Place Recognition}

\author[1,2]{Geoffroy Keime \orcidlink{0009-0000-3451-9458}}
\author[1,3]{Nicolas Cuperlier \orcidlink{0000-0001-7610-3143}}
\author[1,2]{Benoit R. Cottereau \orcidlink{0000-0002-2624-7680}}

\affil[1]{
IPAL, CNRS IRL 2955, Singapore}
\affil[2]{Univ Toulouse, CNRS, CerCo UMR 5549, Toulouse, France}
\affil[3]{Laboratoire ETIS UMR 8051, CY Cergy-Paris Université, ENSEA, CNRS, Cergy, France}

\date{}  

\maketitle

\begin{abstract}
Reliable visual place recognition (VPR) under dynamic real-world conditions is critical for autonomous robots, yet conventional deep networks remain limited by high computational and energy demands. Inspired by the mammalian navigation system, we introduce SpikeVPR, a bio-inspired and neuromorphic approach combining event-based cameras with spiking neural networks (SNNs) to generate compact, invariant place descriptors from few exemplars, achieving robust recognition under extreme changes in illumination, viewpoint, and appearance. SpikeVPR is trained end-to-end using surrogate gradient learning and incorporates EventDilation, a novel augmentation strategy enhancing robustness to speed and temporal variations. Evaluated on two challenging benchmarks (Brisbane-Event-VPR and NSAVP), SpikeVPR achieves performance comparable to state-of-the-art deep networks while using 50× fewer parameters and consuming 30–250× less energy, enabling real-time deployment on mobile and neuromorphic platforms. These results demonstrate that spike-based coding offers an efficient pathway toward robust VPR in complex, changing environments.

\end{abstract}


\section*{Introduction} \label{sec:introduction}

Visual Place Recognition (VPR) is an essential task for navigation, aiming to identify a previously visited physical location based solely on visual input. It consists of matching a query image against a database of known places despite substantial variations in appearance. Robust VPR is critical for a wide range of technologies, including augmented reality, autonomous driving, and mobile robotic platforms such as drones. In particular, it plays a key role in loop-closure detection in simultaneous localization and mapping (SLAM) and supports reliable map-based navigation in large-scale indoor and outdoor environments. A central challenge in VPR is achieving reliable recognition under significant environmental variability. Real-world scenes evolve continuously due to changes in illumination, weather, and season, as well as variations in viewpoint. In addition, occlusions and dynamic elements, such as pedestrians or vehicles, can further alter the visual content of a scene, making consistent place recognition particularly difficult (see Supplementary Figures \ref{fig:annex1} and \ref{fig:annex2} for some illustrative examples). Nevertheless, recognition systems must consistently and efficiently identify locations across these variations. This difficulty is further exacerbated by the intrinsic visual redundancy of natural environments: distinct places often share similar structures and textures, a phenomenon known as perceptual aliasing \cite{vprtuto2024}. Overcoming this ambiguity requires learning representations that capture the enduring identity of a place rather than its transient visual appearance.
This challenge is further compounded by limited supervision, as only a small number of examples per location are typically available for training.

State-of-the-art VPR methods use deep neural networks to encode RGB images into compact, invariant descriptors (e.g., NetVLAD \cite{arandjelovicNetVLADCNNArchitecture}), which are then matched against a database to recognize previously visited locations (see Figure \ref{fig:1} and \cite{masoneSurveyDeepVisual2021a} for a review). While highly effective, these approaches remain difficult to deploy on resource-constrained platforms such as mobile robots or embedded devices, as the inference latency and energy consumption associated with their tens of millions of real-valued parameters render them impractical for portable implementations.

In contrast, animals such as rodents and primates, including humans, can perform rapid and robust  place recognition \cite{peelen2009, Mallot2018}, even under challenging conditions such as revisiting a location from an opposite viewpoint. This ability relies on specialized neural circuits for visual navigation, notably the entorhinal cortex (EC) and the hippocampus. In the hippocampus, neurons known as place cells become active when an animal occupies or recognizes a specific location, effectively encoding spatial memory \cite{okeefeGeometricDeterminantsPlace1996a, rolls2023}. The entorhinal cortex acts as a critical gateway to place cells, integrating and compressing sensory information, particularly visual input, from across the neocortex \cite{cantoWhatDoesAnatomical2008, brunetalscience02} to supply them with the relevant spatio-temporal features necessary for spatial representation \cite{eichenbaumIntegrationSpaceTime2017}. These mechanisms are highly energy-efficient, with visual information transmitted from the retina to the entorhinal cortex and hippocampus through sparse, spike-based codes \cite{Pitkow2012, wang2019} (see Figure \ref{fig:1}, second row). Remarkably, the human brain requires only about 20 watts to sustain such complex processing \cite{mink1991}. Emulating these principles in artificial systems could therefore enable frugal VPR systems with competitive real-time performance while drastically reducing inference costs, thereby facilitating deployment on highly constrained embedded hardware.

Here, we introduce SpikeVPR, a bio-inspired and frugal VPR system that extracts invariant place descriptors from data captured by an event-based camera and processed with a spiking neural network (SNN) (see Figure \ref{fig:1}, third row). Unlike conventional synchronous cameras, which transmit luminance or RGB values for all pixels at a fixed frequency, event-based cameras operate like the retina, sending binary spikes only when and where luminance changes occur. This drastically reduces the amount of information to be processed. Event-based cameras are robust to lighting variations and motion blur \cite{lichtsteiner128times1281202008}, making them well suited for encoding visual information under diverse environmental conditions, including day/night cycles, weather, or seasonal changes. SNNs can process these events efficiently with low energy consumption, thanks to their spike-based coding. While a few VPR approaches have either combined event data with deep neural networks \cite{fischerEventbasedVisualPlace2020a, leeEventVLADVisualPlace2021, leeEvReconNetVisualPlace2023}, or processed RGB frames using SNNs by encoding image intensities into spike trains \cite{hussainiSpikingNeuralNetworks2022a, hinesVPRTempoFastTemporally2024}, SpikeVPR is, to our knowledge, the first system fully based on SNNs trained end-to-end with surrogate gradient learning on event-based data. We validate its effectiveness through an extensive evaluation on two recent and challenging datasets: the Brisbane-Event-VPR \cite{fischerEventbasedVisualPlace2020a} and NSAVP \cite{carmichaelDatasetBenchmarkNovel2024}. Our results show that SpikeVPR achieves performance comparable to state-of-the-art deep neural network approaches while using 50× fewer parameters and consuming 30× to 250× less energy. Specifically, our contributions are as follows:

\begin{itemize}
    \item We introduce SpikeVPR, the first lightweight, bio-inspired SNN trained end-to-end with surrogate gradient learning on event-based data for visual place recognition.
    \item We propose EventDilation, a novel data augmentation strategy for event-based data that enhances robustness by embedding temporal variations directly into the event representation.
    \item The performance of our approach is extensively evaluated across diverse environments, including on a new dataset (NSAVP), where we provide the first reported results.
    \item We demonstrate that SpikeVPR achieves competitive performance with a remarkably small number of parameters and low energy consumption, making it well-suited for neuromorphic implementation.
\end{itemize}

The complete source code for SpikeVPR is available at: \url{https://github.com/GeoffroyK/SpikeVPR}

\begin{figure}[H]
    \centering
    \includegraphics[width=\textwidth]{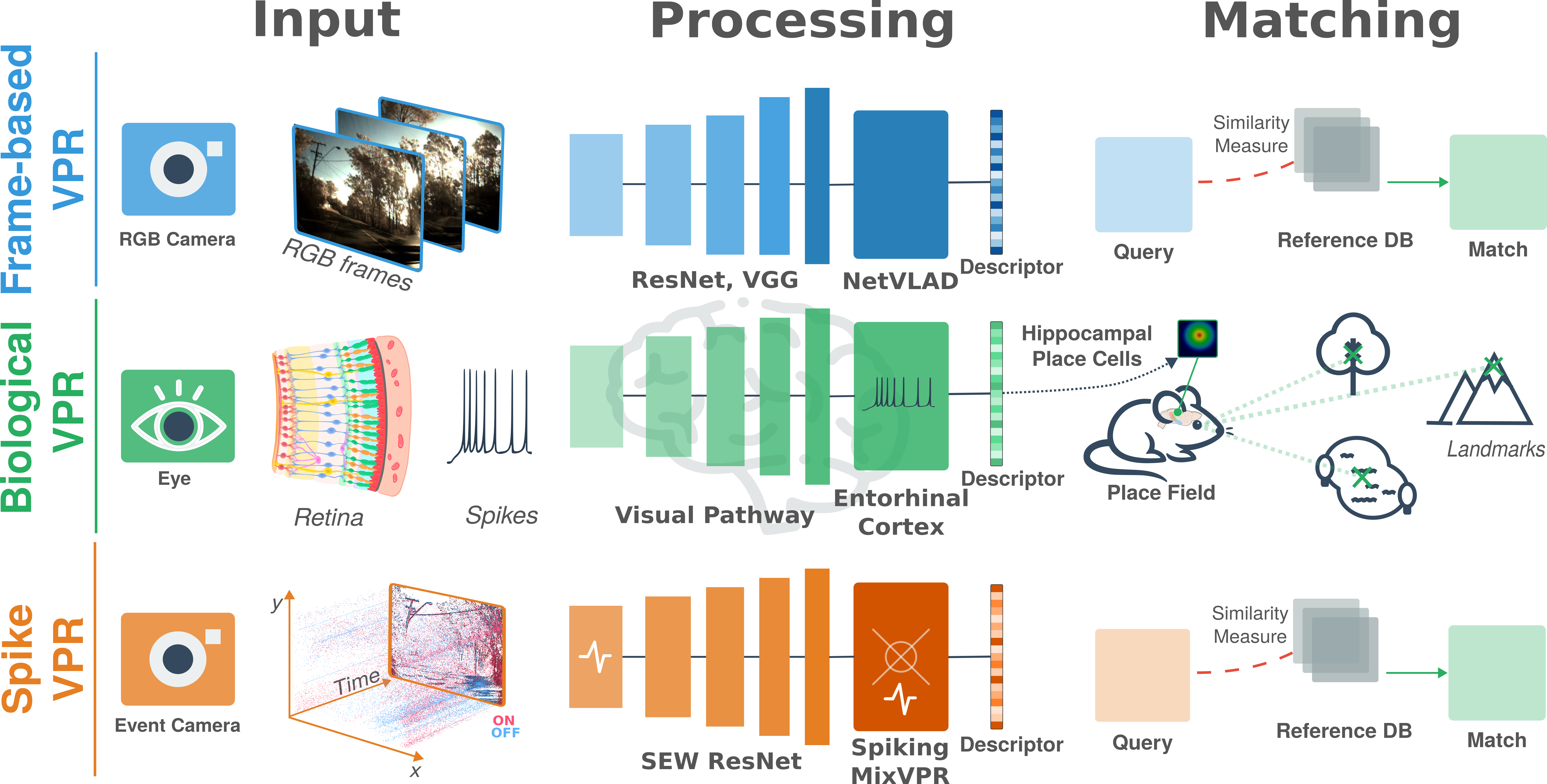}
    \caption{Visual place recognition (VPR) in classical frame-based, biological, and bio-inspired systems. Classical systems (illustrated in blue in the first row) typically rely on RGB images captured at a fixed sampling rate (e.g., 30 or 60 Hz). These images are processed by deep neural networks, such as ResNet or VGG, to extract discriminative descriptors of the different locations (middle panel). The descriptor of a query image is then compared with those stored in memory using a similarity metric to identify the closest match. Although this approach achieves strong retrieval performance, it remains challenging to deploy on resource-constrained platforms because it depends on tens of millions of real-valued parameters, leading to high computational and memory demands that limit their practicality for portable implementations. In contrast, visual place recognition in biological systems (shown in green in the second row) is far more efficient. The retina primarily transmits sparse information as spikes, which occur mostly when changes in illumination, either increments or decrements, are detected in the visual scene. Because spikes are all-or-none signals, their processing along the visual pathway is highly efficient, and the entire system is estimated to consume only around twenty watts.
    At the end of this processing, cortical structures such as the entorhinal cortex encode environmental features that support the formation of hippocampal place cells, neurons that represent specific locations in the explored environment. The inset above the rat illustrates the place field of one such neuron.
Our proposed approach, SpikeVPR (shown in orange in the last row), draws direct inspiration from biological systems. It employs an event-based camera, which, like the biological retina, detects changes in illumination in near real-time. The resulting ‘on’ and ‘off’ spikes are processed by a spiking neural network (SNN) that, similar to the brain, encodes scene descriptors using only binary values. In our implementation, the SNN is built on a SEW ResNet architecture, and scene descriptors are extracted using Spiking MixVPR (see the Materials and Methods).
}
\label{fig:1}
\end{figure}

\section*{Results} \label{sec:results}

\begin{figure}[H]
    \centering
    \includegraphics[width=\textwidth]{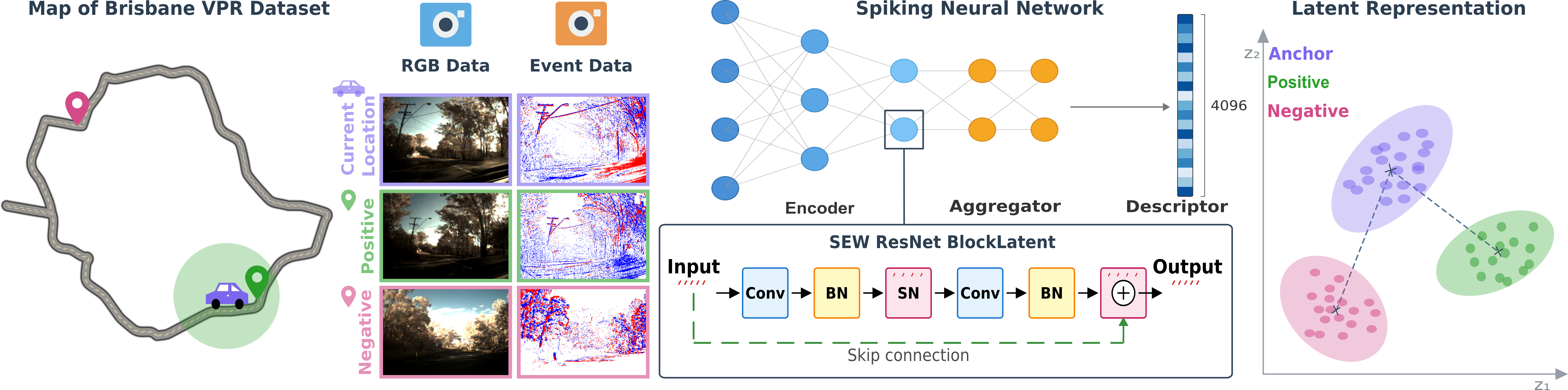}
    \caption{Overview of the SpikeVPR architecture and its training procedure, illustrated using the Brisbane VPR dataset. Given the event stream corresponding to the vehicle’s current location (purple), SpikeVPR generates a 4096-dimensional descriptor through a spiking neural network (SNN). The architecture consists of an encoder built from Spike-Element-Wise (SEW) ResNet blocks (shown in the lower part of the middle panel) followed by an aggregation module. The SNN is trained using surrogate gradient learning (SGL) with a contrastive loss. This objective encourages high similarity between the latent representation of the current location and those of positive examples (i.e., places located within 30 meters of the query and sampled across different traversals, in green), while reducing similarity with descriptors corresponding to negative locations (i.e., other places in pink). RGB images are displayed for visualization purposes only and are not used in the processing pipeline.
}
    \label{fig:2}
\end{figure}

In this work, we draw inspiration from biology to introduce SpikeVPR, a lightweight, neuromorphic-compatible system for recognizing locations that have been visited only a few times, using solely visual input and without relying on GPS, maps, or other external sensors (see Figure \ref{fig:1}). In contrast to conventional deep neural network-based approaches, which often rely on tens of millions of parameters and substantial computational resources, SpikeVPR is explicitly designed for efficiency and deployment on resource-constrained embedded platforms. By leveraging event-based camera data and processing it with spiking neural networks that can learn from only a few exemplars using a contrastive loss optimized via surrogate gradient learning (see Figure \ref{fig:2}), it achieves both low computational overhead and high retrieval accuracy. In the following, We provide a comprehensive evaluation of its performance under various environmental and operational conditions.

\subsection*{Recognition performance in peri-urban environments}\label{sec:results_brisbane}
 
We evaluated SpikeVPR on the Brisbane-Event-VPR dataset \cite{fischerEventbasedVisualPlace2020a}, which comprises six traverses of the same eight-kilometer route captured under varying illumination and traffic density conditions (see the Methods). We adopted the standard geographical tolerance threshold of $\Theta = 30$ meters to define correct matches, consistent with previous VPR work on this dataset \cite{fischerEventbasedVisualPlace2020a, kongEventVPREndtoEndWeakly2022, fischerHowManyEvents2022a}. Under these conditions, the environment contains a total of 578 distinct places. As in previous deep learning approaches for VPR \cite{fischerEventbasedVisualPlace2020a, kongEventVPREndtoEndWeakly2022}, our network was pretrained on a separate dataset to improve representation quality and downstream performance \cite{chenSimpleFrameworkContrastive2020, newell_how_2020}. Specifically, we used two traverses from the NSAVP dataset for pretraining. 

\begin{figure}[H]
    \centering
    \includegraphics[width=\textwidth]{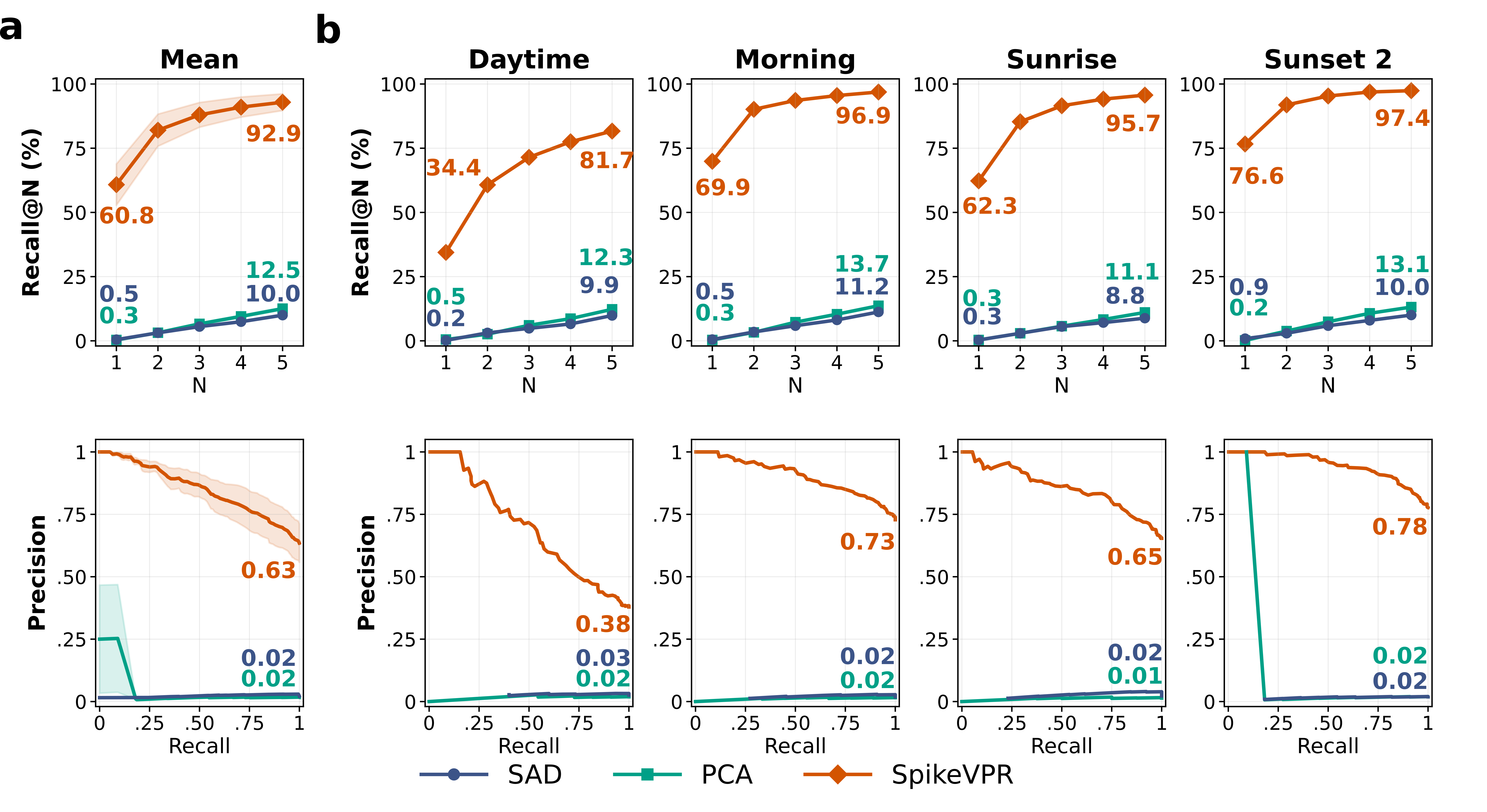}
    \caption{Recall@N (top row) and Precision (bottom row) curves on the Brisbane VPR dataset.
Results obtained with SpikeVPR are shown in orange. For comparison, we include curves estimated using the sum of absolute differences (SAD) method (blue) and principal component analysis (PCA) (green). The values displayed on the left of the first row correspond to the Recall@1. The values displayed on the right correspond to the Recall@5 (first row) and precision Recall@100 (second row). (\textbf{a}) Average performance across all traverses (Daytime, Morning, Sunrise, Sunset 2). Shaded regions indicate the standard error of the mean. (\textbf{b}) Performance measured separately for each tested traverse.}
    \label{fig:3}
\end{figure}

Figure \ref{fig:3} shows the Recall@N performance (the percentage of query locations for which the correct match appears among the top N retrieved candidates from the reference database Sunset 1) and the precision curves  obtained for different traverses (Daytime, Morning , Sunrise and Sunset 2) using SpikeVPR (orange curve). 
The proposed method achieves an average Recall@1 of 60.8\%, with 34.4\% on Daytime, 69.9\%  on Morning, 62.3\% on Sunrise and 76.6\% on Sunset 2 (Figure~\ref{fig:3}, first row, orange values). Considering more candidates further improves retrieval performance: the average Recall@5 reaches 92.9\%, including 81.9\% on Daytime, 96.9\% on Morning, 95.7\% on Sunrise, and 97.4\% on Sunset 2 (Figure~\ref{fig:3}, orange values). This performance is notable given that the network was trained on only three traverses and that the chance level in this setting is approximately $1.7 \times 10^{-3}$. Precision curves further confirm the retrieval quality, with an average Recall@100 of 0.63 (0.38 Daytime, 0.73 Morning, 0.65 Sunrise, and 0.78 Sunset 2; Figure~\ref{fig:3}, second row). To illustrate SpikeVPR’s predictions on this dataset, we provide examples in Supplementary Figure \ref{fig:annex3}. Supplementary Figure \ref{fig:annex4} shows the performance of a network that was not pretrained.

\subsection*{Comparison with state-of-the-art methods}\label{sec:comparisons}

To date, only a limited number of studies have explored the use of spiking neural networks (SNNs) for VPR with spike trains derived from frame-based images (\cite{hussainiSpikingNeuralNetworks2022a, hussainiEnsemble2022, hinesVPRTempoFastTemporally2024, hinesCompactNeuromorphicSystem2025, hussainiApplications}). In these works, performance gains over classical methods such as sum-of-absolute-differences (SAD) were marginal (see, e.g., Table 1 and Figure 2 in \cite{hussainiEnsemble2022}). Because the proposed approaches do not scale to the higher-resolution data considered in the present study, for completeness and fair comparison, we report Recall@N and precision results obtained using SAD (and also a principal component analysis, PCA) under our experimental conditions. On average, SpikeVPR outperforms these baselines by a factor of six (see Figure \ref{fig:3}), an order of magnitude higher than observed for the aforementioned approaches.

\begin{figure}[H]
    \centering
    \includegraphics[width=\textwidth ]{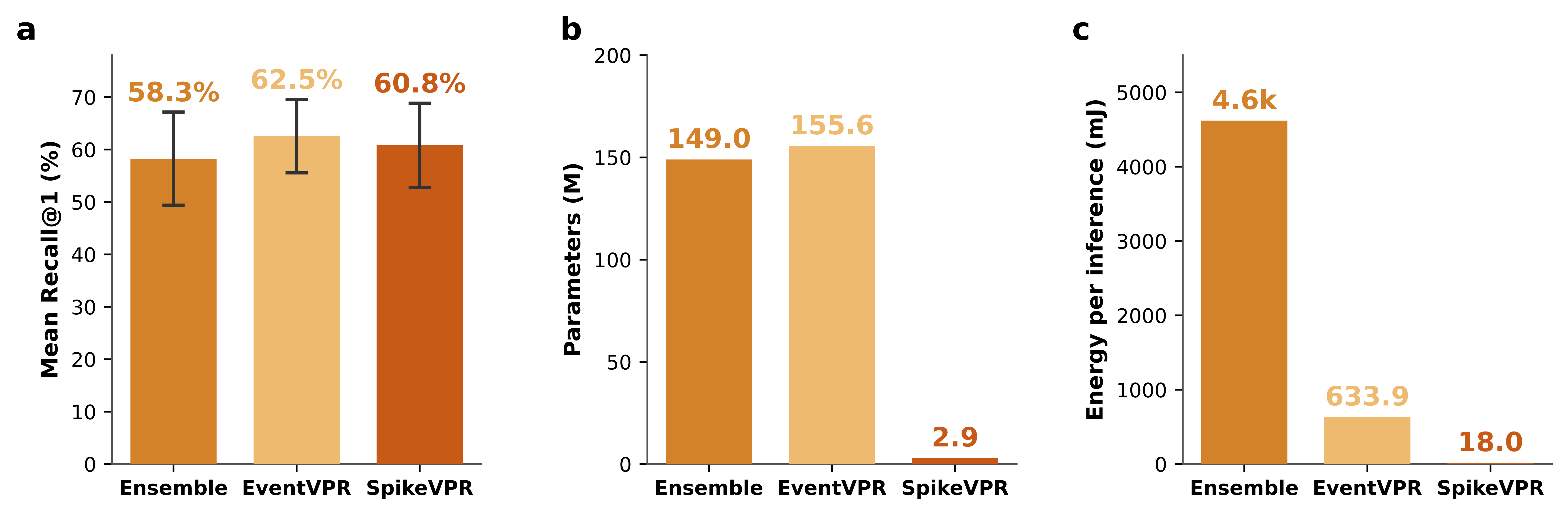}
    \caption{Performance and model complexity of SpikeVPR (dark orange) compared with two recent event-based deep learning approaches: Ensemble (orange) and EventVPR (light orange). (\textbf{a}) Average Recall@1 across the different traverses of the Brisbane-Event-VPR dataset (see Figure~\ref{fig:3} for details). Error bars indicate the standard error computed across traverses. (\textbf{b}) Number of trainable parameters for each method. (\textbf{c}) Estimation of energy consumption of both ANNs (Ensemble, EventVPR) vs SpikeVPR (see Section~\ref{sec:mm_energy}).}
    \label{fig:4}
\end{figure}

To complete our comparisons, Figure~\ref{fig:4}‑a reports the average Recall@1 performance across the various tests shown in Figure~\ref{fig:3} for SpikeVPR and two alternative event-based approaches that do not rely on SNNs: Ensemble and EventVPR. It is important to note that the values for EventVPR were directly extracted from the original publication, as the code to reproduce this method is not publicly available. In contrast, analyses for Ensemble were conducted using the publicly available code at the following GitHub repository: \url{https://github.com/Tobias-Fischer/ensemble-event-vpr}. We observe that Ensemble and EventVPR achieve performances of 58.26 +/- 17.77\% and 62.5 +/- 13.98\%, respectively, which is similar to what is obtained with SpikeVPR (60.8 +/- 16.06\%). However, when comparing the number of parameters across the different approaches (Figure~\ref{fig:4}-b), SpikeVPR uses approximately fifty times fewer parameters. Its energy consumption is also significantly lower, at only 18 mJ per inference, about 250 times less than Ensemble and 30 times less than EventVPR (Figure~\ref{fig:4}-c).Taken together, these results demonstrate that SpikeVPR is substantially more efficient than its competitors while having comparable recognition performance. In terms of computation time, with GPU acceleration, SpikeVPR processes a single input in approximately $9.5$ ms in evaluation mode with gradients disabled, corresponding to over 100 frames per second and demonstrating that the model is well-suited for real-time applications.

\subsection*{Recognition performance in urban traffic environments} \label{sec:results_nsavp}

\begin{figure}[H]
    \centering
    \includegraphics[width=0.8\columnwidth ]{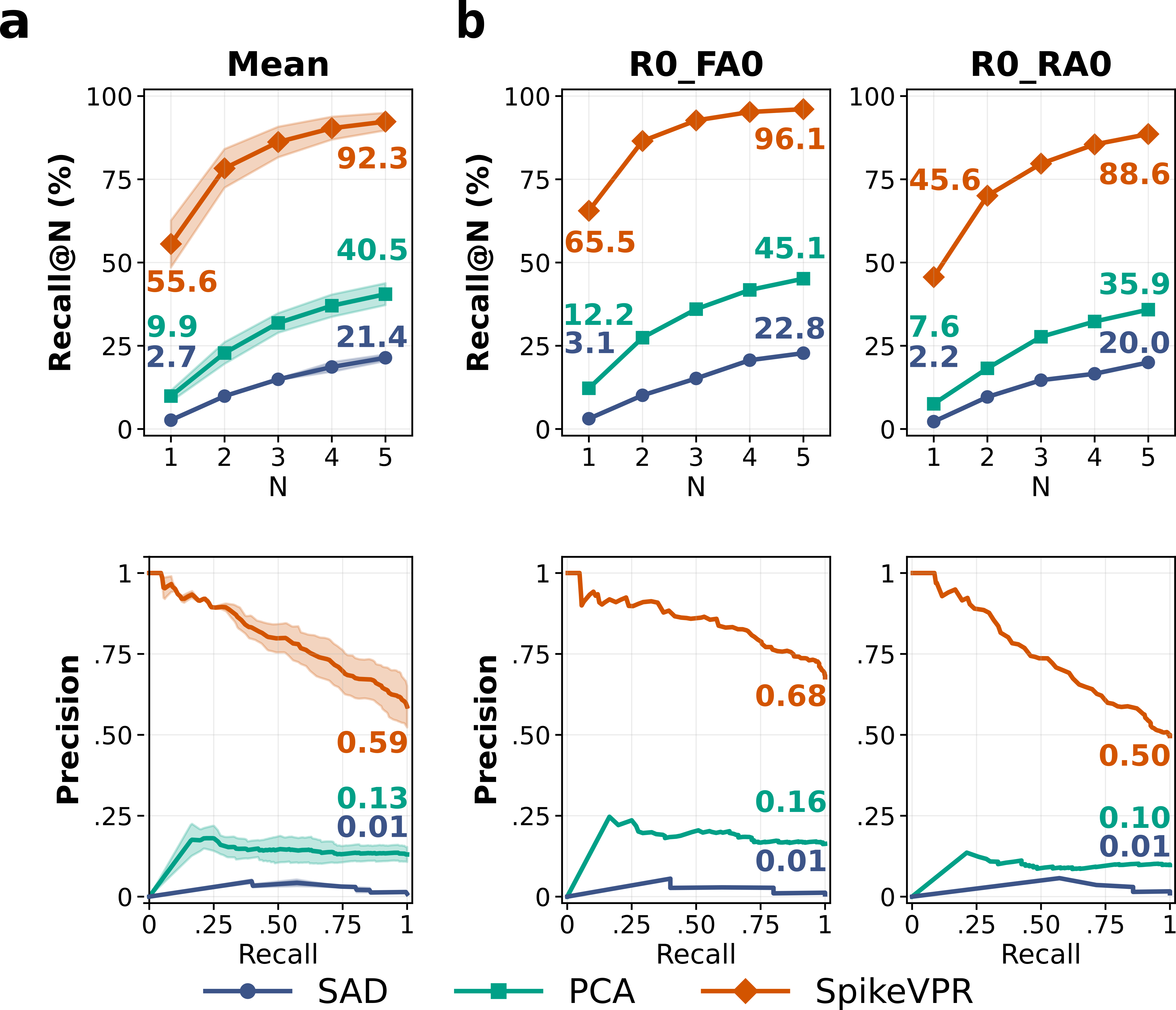}
    \caption{Recall@N (top row) and Precision (bottom row) curves on the NSAVP dataset on Road 0 going forward and on reverse. The references in respective traverses are R0\_FS0 and R0\_RS0. 
    (\textbf{a}) Average performance across all traverses (R0\_FA0 and R0\_RA0). Shaded regions indicate the standard error of the mean. (\textbf{b}) Performance measured separately for each tested traverse. See Figure 3 for more details.}
    \label{fig:5}
\end{figure}

We also train and evaluate our model on the Novel Sensors for Autonomous Vehicle Perception (NSAVP) \cite{carmichaelDatasetBenchmarkNovel2024} dataset.  The dataset was used for event VPR in \cite{josephEnsembleBasedEventCamera2025} following a similar approach as \cite{fischerEventbasedVisualPlace2020a}. To the best of our knowledge, no prior work has reported training on this dataset for event-based VPR. Therefore, direct comparison with existing methods is not possible. We train our network on daytime scenarios (R0 onward, with 1046 places) and evaluate it on the same road in reverse (with 1286 places), which approximates a zero-shot setting. Notably, unlike previous tests performed on the Brisbane dataset, both the encoder and the aggregator of SpikeVPR are trained from random weight initialization for NSAVP. In this setting, the model is able to recognize places from views that were not seen during training, demonstrating the model’s learning capacity to generalize and build upon its learned features. The performance of SpikeVPR on this dataset is presented in Figure~\ref{fig:5}. On average, it achieves a Recall@1 of 55.6, with 65.5\% on R0FA0 and 45.6\% on R0RA0. This corresponds to a 20-fold improvement over SAD and a 5.6-fold improvement over PCA (see the numbers in orange on the left part of each plot on the first row). The same effects were observed when considering the Recall@5 (see the numbers in orange on the right part of each plot on the first row). Differences on precision were even more marked with average values of 0.59 for Spike VPR (0.68 for R0FA0 and 0.5 for R0RA0), 0.01 for SAD and 0.13 for PCA. This corresponds to a 50-fold improvement over SAD and a 4.5-fold improvement over PCA. It is important to emphasize that this retrieval performance was achieved with places observed from opposite viewpoints, a scenario that poses a particularly challenging problem in visual place recognition. To illustrate SpikeVPR’s predictions on this dataset, we provide examples in Supplementary Figure \ref{fig:annex4}.

\subsection*{Impact of Event Dilation on performance} \label{sec:ablation_study}

To better understand the key elements that enable SpikeVPR to retrieve different places, we conducted an ablation study on the data augmentation techniques, including EventDrop \cite{guEventDropDataAugmentation2021}, X-axis reversal, and our proposed augmentation method, event dilation. The model was trained on the NSAVP dataset (Road 0, forward traversal) and evaluated on the reverse traversal of the same road. Table 1 shows the performance (recall@1 and precision) obtained in this case. 

\begin{table}[h]
\centering
\caption{\textbf{Effect of data augmentations on SpikeVPR performance.} 
Recall at rank 1 (R@1, \%) and Area under the precision-recall (PR) curve are reported under different augmentation settings. 
D: event dilation; X: x-axis transform; E: EventDrop.}
\label{tab:augmentation_ablation}
\begin{tabular}{l c c}
\hline
Augmentation & R@1 (\%) & Precision@100 (\%) \\
\hline
None & 14.22 & 14.81 \\
D & 37.78 & 42.81 \\
X & 18.96 & 22.22 \\
E & 20.59 & 22.96 \\
X + E & 24.74 & 27.26 \\
D + E & 37.93 & 45.19 \\
D + X & 45.63 & 49.63 \\
D + X + E & 45.19 & 52.30 \\
\hline
\end{tabular}
\end{table}

Across all experimental configurations of reported in table~\ref{tab:augmentation_ablation}, EventDilation consistently improves network performance. Notably, the Recall@1 and Precision@100 achieved using only event dilation (37,78\% and 42,81\%) outperform those obtained with the two other augmentation techniques (23,74\% and 27.26\%). Under naturalistic conditions, as in the two datasets considered in this work, vehicle speed may vary from one traverse to another. This variation can affect the number of events associated with a given place: for a fixed event histogram duration, the same location captured at a higher vehicle speed will generate more events. By varying the event histogram duration, our Event Dilation augmentation enhances robustness to speed variations, thereby improving VPR performance.

\section*{Discussion} \label{sec:discussion}

In this work, we introduce SpikeVPR, a bio-inspired and frugal visual place recognition (VPR) system that extracts place descriptors from event-based data using a spiking neural network (see Figure \ref{fig:1}). Place retrieval is performed via a few-shot learning strategy based on a contrastive loss function and optimized with surrogate gradient learning (see Figure \ref{fig:2}). SpikeVPR delivers high capabilities while using significantly less parameters than other existing methods. To our knowledge, it is the first fully neuromorphic-compatible approach to combine both event-based data and SNN for non sequential VPR. Nonetheless, several previous studies investigated VPR using either SNNs or event based-data independently.

\subsection*{Comparison with previous SNN approaches}\label{sec:comp_prev_app_SNN}
The growing interest in neuromorphic computing has led researchers to explore SNNs for VPR tasks. Hussaini \textit{et al.} \cite{hussainiSpikingNeuralNetworks2022a} were the first to introduce SNNs for VPR. Their architecture consisted of two-layer network composed of leaky integrate-and-fire (LIF) neurons, trained using a local learning rule based on spike-time-dependent plasticity (STDP). Inputs spikes were generated from RGB images converted into Poisson-distributed spike trains, where pixel intensities were mapped to firing rates. This work was later extended by the same group in  \cite{hussainiEnsemble2022}
with Ensemble SNN, a modular architecture composed of compact and localized SNNs that enabled the system to learn a larger number of places. Although these approaches are promising for robotic applications, particularly because they can be deployed on neuromorphic hardware, they rely on heavily downsampled inputs (images resized to $28 \times 28$ pixels across multiple datasets). Moreover, their performance remains limited. For instance, the precision at 100\% Recall reported for Ensemble SNN is only 7.5 percents higher than that obtained with the sum-of-absolute differences (SAD) baseline (52.6 \% vs 45.1\%; see their Table 1). In sharp contrast, SpikeVPR achieves substantially better performance relative to SAD with its precision at 100\% Recall exceeding SAD by more than thirtyfold ($\times 31.5$) on average on the Brisbane-Event VPR dataset ($346 \times 268$ pixels; see Figure \ref{fig:3}). We attribute this substantial improvement to the use of a contrastive learning rule optimized with surrogate gradients, rather than an unsupervised and local rule such as STDP. By directly optimizing a global objective across the entire network, Spike VPR overcomes the limitations of relying solely on input correlations.

\subsection*{Comparison with previous event-based approaches}\label{sec:comp_prev_app_event}

Several studies have proposed VPR approaches that leverage event-based data in combination with deep neural networks. For example, Fischer et al. \cite{fischerEventbasedVisualPlace2020a} used E2VID \cite{rebecqHighSpeedHigh2021}
(Event-to-Video) to reconstruct frames from event streams, experimenting with different parameters such as the number of events and the duration of the integration window. These reconstructed frames are then fed into a pretrained NetVLAD module, which extracts discriminative features for the resulting RGB images or edge reconstucted frames \cite{fischerEventbasedVisualPlace2020a, leeEventVLADVisualPlace2021, leeEvReconNetVisualPlace2023,josephEnsembleBasedEventCamera2025}. While these approaches achieve strong performance, they cannot be directly implemented on neuromorphic hardware because they rely on real-valued computations. Moreover, their large number of parameters limits their suitability for real-time embedded systems. For instance, Ensemble contains 149 million parameters, excluding the E2VID module, and EventVPR has an estimated 155.6 million parameters. In contrast, SpikeVPR requires only 2.9 million parameters, about fifty times fewer, while achieving comparable performance and operating in real time with hardware acceleration (under 10 milliseconds per input) (see Figure \ref{fig:4}). These results suggest that competitive VPR does not necessarily require increasingly large models. Instead, carefully designed architectures following a frugal AI approach can significantly reduce computational and parameter costs, highlighting the potential of biologically inspired principles for building efficient vision systems.

\subsection*{Biological Plausibility and Neurophysiological Implications}\label{sec:discuss-bio}

SpikeVPR is directly inspired by the mammalian navigation system. By processing event-camera data through a spiking feature extractor, the model emulates the hierarchical processing of the visual cortex using sparse binary codes. This allows for rapid and robust extraction of visual place properties, closely mirroring those observed in human behavioral studies \cite{Mononenetal2025JN, peelen2009, Mallot2018}. Remarkably, SpikeVPR demonstrates view invariance, maintaining robust performance even when places are observed from opposite viewpoints (as shown on the NSVAP dataset), thereby capturing a key aspect of human spatial cognition \cite{king2002,Segenetal2021}.

At the neural level, SpikeVPR generates descriptor vectors analogous to the spatial codes observed in brain regions such as the entorhinal cortex (EC) that notably provide the primary cortical input to hippocampal place cells. Although this cortical area is well known for its grid cells, these constitute only a minority of its neural units. In fact, most of its cells exhibit spatial responses and belong to other neural populations (approximately 78\%, see \cite{diehlGridNongridCells2017}). In our model, the MixVPR aggregator acts as a high-level neural encoder of visual environmental features. The resulting spatial code supports the formation of place-cell–like responses, exhibiting output signatures that globally approximate the spatial activity of these multiple entorhinal neural populations. Furthermore, whereas place cells can rapidly form stable place fields in novel environments \cite{moserOneShotMemoryHippocampal2003}, neurons in the entorhinal cortex gradually encode more general features that remain stable over time \cite{mcclelland_why_1995,schapiro_complementary_2017}. This distinction is consistent with the iterative learning process implemented in our model through the surrogate gradient rule. Altogether, this close alignment with behavioral and neurobiological data suggests that SpikeVPR could serve as an in silico model of VPR across different mammalian species. For instance, neural properties across the network’s layers could be characterized using recent explainable AI approaches \cite{ranconTemporalRecurrenceGeneral2025} and compared with brain recordings obtained during a variety of navigation tasks \cite{schrimpfBrainScoreWhichArtificial2018}, or used to generate predictions for future experiments. Pursuing this avenue represents a key direction for our group in the near future.

\subsection*{Implementation on neuromorphic hardware} \label{sec:implementation_neuro}

SpikeVPR is a feedforward and stateless SNN, as neuron membrane potentials are reset after processing each place (i.e., each event histogram). While stateful architectures, such as those based on leaky integrate-and-fire neurons, may be an interesting alternative for certain visual navigation tasks (e.g., sequential VPR), they typically rely on recurrent operations that are difficult to deploy on dedicated neuromorphic hardware. Thanks to its simple architecture, SpikeVPR can be directly mapped onto dedicated chips such as Intel Loihi \cite{daviesLoihiNeuromorphicManycore2018}, IBM TrueNorth \cite{akopyanTrueNorthDesignTool2015}, or BrainChip’s Akida \cite{vanarseHardwareDeployableNeuromorphicSolution2019}, which are designed to exploit sparse binary spike tensors in SNNs. Due to the summation operations performed at the end of its residual layers, SpikeVPR may produce integer-valued spike counts rather than strictly binary spikes. This does not pose a practical limitation: in most digital neuromorphic chips, spikes are transmitted as multi-bit messages that include source and/or destination addressing along with a small payload encoding graded spike values (see, e.g., \cite{daviesLoihiNeuromorphicManycore2018}). In our case, spike counts lie in the range [0, 2] and can therefore be encoded using only 2 bits. For hardware that supports only binary spikes, a spike count of (N) can be implemented as (N) sequential binary spike events.

From a broader perspective, SpikeVPR could be deployed on neuromorphic hardware alongside other recent event-based SNN approaches for navigation-related visual tasks. These include depth estimation \cite{ranconStereoSpikeDepthLearning2022} (see \cite{brito235TOPSDepthwise} for an example of hardware implementation), optical flow computation \cite{cuadradoOpticalFlowEstimation2023a}, as well as visual odometry and SLAM. Together, these methods could enable highly energy-efficient visual navigation systems for mobile platforms while providing redundancy alongside complementary localization systems, thereby improving overall fault tolerance. This capability is particularly relevant for self-driving cars, autonomous mobile robots, and especially unmanned aerial vehicles, where real-time performance and low power consumption are critical constraints. Beyond standard navigation, SpikeVPR could support a range of high-impact applications, from assistive spatial awareness for visually impaired individuals to industrial-grade structural change detection for infrastructure monitoring.

\section*{Methods}\label{sec:materials_and_methods}

We used PyTorch and SpikingJelly \cite{fangSpikingJellyOpensourceMachine2023a} as our primary development frameworks. PyTorch is one of the most widely adopted libraries for deep learning and automatic differentiation, while SpikingJelly, an open-source framework for spiking neural networks built on top of PyTorch, has seen steadily growing popularity in recent years.

\subsection*{Datasets} \label{sec:nav_datasets}
We trained and tested our model (SpikeVPR) on two event based VPR dataset, the Brisbane-VPR dataset \cite{fischerEventbasedVisualPlace2020a} and the Novel Sensors for Autonomous Vehicle Perception (NSAVP) \cite{carmichaelDatasetBenchmarkNovel2024} dataset. For both of them, the event-based camera is mounted directly on the vehicle’s windshield and records multiple traverses of the same routes, enabling the capture of identical physical locations under varying conditions. Ground truth is obtained using GPS and an Inertial Measurement Unit (IMU), which together provide the vehicle’s 6-DoF pose. The Brisbane dataset was recorded using a DAVIS346 camera with a resolution of $346 \times 268$, whereas the NSAVP dataset was acquired with a DVXplorer camera featuring a spatial resolution of $640 \times 480$ pixels.

For the Brisbane dataset, following established protocols \cite{fischerEventbasedVisualPlace2020a, kongEventVPREndtoEndWeakly2022}, we exclude the night traverse from both evaluation and training due to severely degraded visual information under the extreme low-light conditions encountered by the DAVIS346 event camera. Thus, we have 5 available traverses that are splitted between training and test sets.

\subsection*{Event representation and preprocessing} \label{sec:event_representation}
Event data consist of a list of asynchronously recorded events $e_i$ taking the following form:

$$
e_i=\left[t_i,x_i,y_i,p_i\right],i\in\left\{1,2,3,...,N\right\},
$$
with t the timestamp of the event, x and y its position and p its polarity ($\pm 1$). To facilitate the processing of event data by deep neural networks, particularly for supervised learning via backpropagation, it is common to aggregate events into structured representations such as event histograms or voxel grids. This event representation can be defined in several ways, for example by fixing the number of events, integrating events over a predefined time window, or using a learnable selection mechanism \cite{gaoEndtoEndBroadLearning2020b}. Ultimately, the choice of representation depends on the specific requirements of the user or application. In our case, following a similar approach to \cite{ranconStereoSpikeDepthLearning2022}, we adopted an event histogram representation. Unlike \cite{gehrigDSECStereoEvent2021a}, we deliberately avoided normalizing the histogram to prevent the introduction of floating-point values at the input stage, thereby reducing computational overhead prior to inference and ensure easier hardware implementation with no floating point values.

\subsection*{Event data augmentation}\label{sec:data_augmentation}
A key challenge in retrieval tasks like VPR in real-world scenarios is the limited amount of training data available per location (or traverse). Unlike standard classification tasks with datasets such as NMNIST or ImageNet, each location (or class) typically has very few examples, often only three traverses for training and two for testing. In addition, many locations along a traverse can be highly similar, further increasing the difficulty of training (see Supplementary Figure \ref{fig:annex1}).

In this context, data augmentation can be an effective way to help data-driven models better capture the underlying training distribution. Although an increasing number of studies have adapted classical augmentation methods to event-based data \cite{guEventDropDataAugmentation2021, shenEventMixEfficientData2023, dongEventZoomProgressiveApproach2024, liNeuromorphicDataAugmentation2022}, no universally accepted standard has yet emerged. In this work, we use three types of augmentation. EventDrop \cite{guEventDropDataAugmentation2021} randomly modifies the event stream along a specific dimension, either spatial or temporal, and has proven effective for deep learning applications, particularly in VPR \cite{kongEventVPREndtoEndWeakly2022}. However, it only removes events and does not fully leverage the variations in event distributions for the same locations across different scenarios. This limitation is especially pronounced in event-based VPR, where lighting changes can cause significant differences in the event patterns captured at the same place. 

To address this issue, we introduced a new augmentation, Event Dilation, that applies variable temporal window integration during training, centered on a specific duration. The length of each training example is randomly selected within fixed thresholds, effectively acting as a regularizer for the network. This not only introduces invariance to the vehicle’s speed but also encourages the model to capture finer event patterns, thereby enhancing the quality of its latent representation of the visual input.

Given an event stream of higher temporal resolution,
\[
E = \{ (x_i, y_i, t_i, p_i) \mid i = 1, \dots, N \},
\]

we sample a random temporal window $\Delta t$ 
$$
\Delta t\sim\mathcal{U}\left(t_{min},t_{max}\right)
$$
and define the dilated event set as:
\begin{equation}
E' = \left\{ (x_i, y_i, t_i, p_i) \in E \,\middle|\, 
t_i \in \left[ t_c - \frac{\Delta t}{2},\, t_c + \frac{\Delta t}{2} \right] \right\},
\label{eq:event_window}
\end{equation}
where $t_c$ denotes the chosen center time, $\mathcal{U}\left(t_{min},t_{max}\right)$ is a uniform distribution between the temporal boundaries and $E'$ represents the dilated event set.

\subsection*{Neuron model} \label{sec:neuron_model}

In this work, we considered the McCulloch and Pitts model \cite{mccullochLogicalCalculusIdeas1943}, which is mathematically equivalent to the Integrate-and-Fire (IF) neuron model without temporal recurrence:

\begin{equation}
\tau_m \frac{dV_i(t)}{dt} = -V_i(t) + R_m I_i(t),
\label{eq:if_dynamics}
\end{equation}

\noindent
where $\tau_m$ is the membrane time constant, $R_m$ is the membrane resistance, and $I_i(t)$ represents the input current.
When the membrane potential reaches the threshold $V_{\text{th}}$, the neuron emits a spike and its potential is reset to $V_{reset}$:

\begin{equation}
\text{if } V_i(t) \geq V_{\text{th}} \Rightarrow s_i(t) = 1,  V_i(t) \leftarrow V_{\text{reset}}
\label{eq:if_spike}
\end{equation}

\noindent
In practice, SNNs are typically simulated in discrete time, discretizing Eq.\ref{eq:if_dynamics} with a time step $\Delta t$ yields:

\begin{equation}
V_i[t+1] = V_i[t] + \frac{\Delta t}{\tau_m}\left(- V_i[t] + I_i[t]\right)
\label{eq:if_discrete}
\end{equation}

\noindent
As the McCulloch and Pitts model is stateless, the neuron does not maintain or integrate its membrane potential over time. Each input is processed in a single forward pass. Derived from \ref{eq:if_discrete}, we formulate our neuron model as:


\begin{equation}
s_i[t] = \Theta(V_i[t] - V_{\text{th}}) \text{, with } V_i = \sum_j {w_{ij}x_j}
\label{eq:heavyside_discrete}
\end{equation}

\noindent Where $\Theta$ denotes the Heavyside step function, $V_i$ the membrane potential of the $i$-th neuron and $V_{th}$ denotes the potential threshold of the neuron.

\subsection*{Network architecture} \label{sec:network_architecture}

\subsubsection*{Feature extractor}
To extract low-level features from the input event frames, we trained a fully spiking encoder from scratch. A common challenge when training SNNs is the vanishing/exploding gradient problem, which can result in null gradients and make backpropagation-based training difficult or even infeasible. To address this issue, we adopted the Spike-Element-Wise ResNet (SEW-ResNet) architecture proposed by Fang et al. \cite{fangDeepResidualLearning2022}. SEW-ResNets adapt traditional spiking ResNets by modifying the residual block structure: instead of applying the combination function before the Heaviside step function, an element-wise binary operation $g$ (either \emph{ADD}, \emph{AND}, or \emph{IAND}) combines the spiking neuron output with the identity mapping \emph{after} the spiking operation (see figure 2). This architectural change effectively mitigates gradient vanishing/explosion during training. The binary function $g$ makes this model suitable for neuromorphic hardware implementation. Moreover, the identity mapping enables the network to scale to deeper architectures, which is essential for extracting hierarchical features from high-dimensional event data. In our implementation, we adopted a stateless configuration with a timestep of $T = 1$ and the \emph{ADD} operation, as this setting demonstrated the best performance in the original work.

\subsubsection*{Depthwise separable convolutions}

To further optimize the encoder for lightweight deployment on neuromorphic hardware, we employed depthwise separable convolutions \cite{cholletXceptionDeepLearning2017} within the SEW-ResNet architecture. Standard convolutions in SNNs can be computationally expensive and parameter-heavy. Following a procedure similar to that in \cite{ranconStereoSpikeDepthLearning2022} for stateless SNNs, this approach drastically reduces the number of floating-point operations per second (FLOPS) compared to their ANN counterparts. In practice, depthwise separable convolutions decompose a standard convolution into two steps: a depthwise convolution, which applies a single filter to each input channel, followed by a pointwise ($1 \times 1$) convolution that combines the resulting outputs. This factorization drastically reduces the number of parameters and is known to be efficient for neural networks deployed on mobile operators like in mobileNet implementations \cite{sandlerMobileNetV2InvertedResiduals2019}.
Furthermore, this approach has demonstrated success, as a similar encoder architecture \cite{ranconStereoSpikeDepthLearning2022} was implemented on simulated neuromorphic hardware in \cite{brito235TOPSDepthwise}.

\subsubsection*{Feature aggregator}

Feature aggregation is commonly used in VPR to combine encoder outputs into a compact representation. Before the deep learning era, these features were typically handcrafted. The introduction of NetVLAD \cite{arandjelovicNetVLADCNNArchitecture} provided a fully differentiable alternative, enabling end-to-end training with feature extractor networks. Subsequent studies have proposed alternative aggregation mechanisms to further improve performance \cite{bertonRethinkingVisualGeolocalization2022,revaudLearningAveragePrecision2019, ali-beyMixVPRFeatureMixing2023, bertonMegaLocOneRetrieval2025}, altough NetVLAD remains widely used, including in event-based VPR \cite{fischerEventbasedVisualPlace2020a,
josephEnsembleBasedEventCamera2025, kongEventVPREndtoEndWeakly2022}. For neuromorphic deployments, however, NetVLAD presents practical limitations. Its soft-assignment mechanism relies on softmax operations, which require floating-point computations and can be inefficient on dedicated hardware. Aggregators based on sequences of linear transformations offer a more hardware-friendly alternative with lower computational cost. In this context, we adopt MixVPR \cite{ali-beyMixVPRFeatureMixing2023}, which avoids floating-point operations and maintains a relatively small parameter footprint by using a series of linear projections that capture complementary spatial information. We adapted the original architecture of MixVPR into a Spiking Architecture, making it neuromorphic compatible.

\subsection*{Contrastive loss function} \label{sec:loss}
Standard deep models for VPR typically rely on contrastive losses to ensure that the network minimizes the distance between similar locations in the latent space while maximizing the separation between unrelated ones. Contrastive learning enforces a form of semi-supervised training that encourages the network to extract and leverage mutual information from the inputs.
In VPR, the most commonly used loss function is the Triplet Margin Loss \cite{schroffFaceNetUnifiedEmbedding2015}, which was notably employed to train NetVLAD \cite{arandjelovicNetVLADCNNArchitecture} and continues to serve as a standard baseline in the literature. Its objective is to minimize the distance in the latent space between an anchor vector and a positive vector, while simultaneously maximizing the distance to a negative example.

However, this loss function has several limitations. First, it does not always push negative examples sufficiently far apart, which can lead to representation collapse, as described in the original paper \cite{schroffFaceNetUnifiedEmbedding2015}, where the network maps all inputs to nearly identical representations. To address this, some works introduced an additional random negative component to the loss, effectively extending the triplet to a quadruplet, which helps to better separate negative representations in the latent space. Additionally, triplet-based approaches require explicit mining strategies to select informative hard negatives, which adds significant computational overhead during training.

We believe that such an approach is suboptimal compared to more modern contrastive learning methods that do not rely on explicit negative mining. Following Chen et al. \cite{chenSimpleFrameworkContrastive2020}, we adopted the Normalized Temperature Cross Entropy (NT-Xent) loss. This loss leverages all elements in the batch to implicitly serve as negative examples, with the number of negatives directly proportional to the batch size.

\begin{equation}
    \mathcal{L}_{i,j} = -\log \frac{\exp(\text{sim}(\mathbf{z}_i, \mathbf{z}_j)/\tau)}{\sum_{k=1}^{2N} \mathds{1}_{[k \neq i]} \exp(\text{sim}(\mathbf{z}_i, \mathbf{z}_k)/\tau)}
    \label{eq:ntxentlossfunc}
\end{equation}

The temperature parameter $\tau$  controls the concentration of the similarity distribution: lower values increase the penalty for hard negatives (similar but distinct examples), while higher values smooth the distribution and treat all negatives more uniformly. In our experiments, we set the temperature parameter $\tau = 0.07$ to penalize hard negatives, as natural scenes often feature highly similar road layouts. Our similarity measurements are based on cosine similarity, which is defined as: 

\begin{equation}
\text{sim}(\mathbf{u}, \mathbf{v}) = \frac{\mathbf{u}^\top \mathbf{v}}{\|\mathbf{u}\|_2 \, \|\mathbf{v}\|_2}
\label{eq:cosine_similarity}
\end{equation}

\subsection*{Training with surrogate gradient learning} \label{sec:training_procedure}

The discontinuous activation function in spiking neuron models poses a fundamental challenge for gradient-based optimization, as its derivative is zero almost everywhere and undefined at threshold. Consequently, standard backpropagation cannot be applied directly. In recent years, however, this limitation has been addressed through surrogate gradient learning \cite{neftciSurrogateGradientLearning2019a}, which substitutes the non-differentiable spike function derivative with a smooth, differentiable approximation during the backward pass while preserving the original function in the forward pass. In our case, we use the Sigmoid surrogate function.
This approach enables the training of SNNs via backpropagation, facilitating the development of deeper architectures that more closely resemble artificial neural networks in both structure and learning dynamics, while preserving the energy-efficiency advantages inherent to spike-based computation.

\subsection*{Experimental setup}
\subsubsection*{Training on Brisbane-Event-VPR}\label{sec:mm_brisbane_training}
Our model was trained following a protocol similar to that described in \cite{kongEventVPREndtoEndWeakly2022}. Among the five available traverses, three were used for training (excluding the night scenario), while the remaining two were reserved for evaluation. The Sunset1 traverse was used exclusively as a reference sequence during testing. The model was optimised using the contrastive loss function described in Equation~\ref{eq:ntxentlossfunc} with a batch size of 64, constrained by hardware limitations. Online data augmentation was applied during training, comprising x-axis reversal, EventDrop and EventDilation, as reported in Table~\ref{tab:augmentation_ablation}. We employed the AdamW optimizer with a learning rate schedule consisting of a warm-up phase followed by cosine annealing. We report in the main manuscript results obtained using a SpikeVPR model pretrained on the forward scenario of Road~0 from the NSAVP dataset. Results obtained without pretraining are shown in Supplementary Figure \ref{fig:annex5}For Recall@N and precision–recall curves, we followed the evaluation protocol of \cite{fischerEventbasedVisualPlace2020a, kongEventVPREndtoEndWeakly2022} to ensure fair comparisons.
\subsubsection*{Training on NSAVP}
In the NSAVP dataset, the visual scene is captured using a DVXplorer, a more recent event camera with a spatial resolution of $640\times
480$ pixels. To enable a direct comparison with results obtained on the Brisbane dataset captured using a DAVIS346 sensor (resolution: $346 \times 260$), we applied the downsampling method proposed in \cite{ghosh_evdownsampling_2025}, which maps the DVXplorer event stream directly to the DAVIS346 resolution. This approach also offers the advantage of being compatible with online, hardware-accelerated execution, making it well-suited for real-time deployment. Following the downsampling of the event stream, SpikeVPR was trained using the same procedure described in the previous section. The resulting models, initially trained on the forward scenario, were subsequently used as pretrained initializations for fine-tuning on Brisbane-Event-VPR.

\subsection*{Evaluation metrics}
To evaluate model performance, we use two metrics commonly employed in VPR: Recall@N and Precision-Recall (PR) curves. Recall@N is computed by retrieving the top N nearest points in the dataset, in the latent space, for each encoded query, and measuring the percentage of correctly identified queries across the entire test set. Formally, Recall@N is defined as:

\begin{equation}
\text{Recall@}N = \frac{1}{|U|} \sum_{u \in U} \mathds{1}\!\left( R_u \cap \text{Top-}N_u \neq \emptyset \right)
\end{equation}

\noindent where $U$ is the set of query images, $R_u$ is the set of ground-truth relevant references for query $u$, $\text{Top-}N_u$ denotes the set of $N$ highest-ranked retrieved candidates for query $u$, and $\mathbb{1}(\cdot)$ is the indicator function returning 1 when the condition is satisfied and 0 otherwise. A retrieved reference is considered correct if its geographic distance to the query location is within 30 meters, consistent with prior studies using this dataset \cite{fischerEventbasedVisualPlace2020a, kongEventVPREndtoEndWeakly2022}.

To complete the evaluation and the performances of our model, we compute the Precision-Recall curve that is defined as:

\begin{equation}
\text{Precision}(\tau) = \frac{TP(\tau)}{TP(\tau) + FP(\tau)}
\end{equation}

\begin{equation}
\text{Recall}(\tau) = \frac{TP(\tau)}{TP(\tau) + FN(\tau)}
\end{equation}

\noindent where $TP(\tau)$, $FP(\tau)$, and $FN(\tau)$ denote the number of true positives, false positives, and false negatives at threshold $\tau$, respectively. The resulting curve captures the trade-off between the system's ability to retrieve all relevant matches and its tendency to avoid false retrievals, providing a more comprehensive view of performance than Recall@$N$ alone.

\subsection*{Energy consumption estimation} \label{sec:mm_energy}

To evaluate the energy efficiency of SpikeVPR, we estimate the energy consumption associated with synaptic operations and memory accesses during inference. We follow the analytical frameworks proposed in \cite{dampfhofferAreSNNsReally2023, lemaireAnalyticalEstimationSpiking2023}, which model the energy of SNNs independently of a specific hardware target. Both studies highlight the importance of considering memory accesses, rather than focusing solely on synaptic operations, when estimating the energy budget of SNNs on digital hardware.

\subsubsection*{Neuron energy model}
As mentioned in equation \ref{eq:heavyside_discrete}, SpikeVPR employs the McCulloch and Pitts model which is referred as the Integrate-and-Fire with instantaneous synapses (IF+inst) throughout the entire pipeline (SEW Resnet + spiking MixVPR). In this model, each incoming spike triggers a single accumulate (AC) operation per activated synapse, with no per-timestep neuron update overhead, as the membrane has no temporal dynamics.

Following Dampfhoffer \textit{et al.}~\cite{dampfhofferAreSNNsReally2023} (Eq. ~4), the energy of an IF+inst layer is:

\begin{equation}
E_{\text{IF+inst}} = N_{\text{syn}} \times N_{\text{spikes/syn}} 
\times \left( E^R_{\text{weight}} + E^R_{\text{state}} 
+ E^W_{\text{state}} + E_{\text{AC}} \right)
\label{eq:dampfhoffer_if_inst}
\end{equation}
where $N_{\text{syn}}$ is the total number of synapses, 
$N_{\text{spikes/syn}}$ is the average number of spikes 
received per synapse per inference, $E^{R/W}$ denotes the 
energy cost of reading or writing from SRAM, and 
$E_{\text{AC}}$ is the energy cost of a single accumulate 
operation. Note that the SEW-ResNet ADD 
connection produces integer-valued outputs $\in \{0, 1, 2\}$ 
at residual junctions by summing two binary spike tensors; 
for the purpose of this analysis, these are treated as 
spike-equivalent operations under the AC model.

\subsubsection*{Layer-wise energy decomposition}

For a finer-grained analysis, we employ the layer-wise decomposition proposed by Lemaire 
\textit{et al.}~\cite{lemaireAnalyticalEstimationSpiking2023} 
(Eq.~18), which breaks down the total energy into three components:
\begin{equation}
E = E_{\text{mem}} + E_{\text{ops}} + E_{\text{addr}}
\label{eq:lemaire_total}
\end{equation}
where $E_{\text{mem}}$ accounts for all SRAM read and write 
accesses (input spikes, weights, membrane potentials, and 
output spikes), $E_{\text{ops}}$ captures the cost of 
synaptic operations (spike integration, bias integration, 
and membrane reset), and $E_{\text{addr}}$ accounts for the 
addressing overhead inherent to sparse event-driven 
computation. This decomposition is applied identically to 
both SNN and ANN models, ensuring a consistent comparison 
framework. Per-layer spike counts ($\theta_l$, the total 
number of spikes emitted by layer~$l$) are measured 
empirically by using the native SpikingJelly spike monitor on each IF neuron 
module and averaging over the different test traverses.

\subsubsection*{ANN baseline estimation}

For comparison with existing VPR models based on ANN, we estimate the 
energy consumption of NetVLAD with VGG-16~\cite{fischerEventbasedVisualPlace2020a} 
and ResNet-34~\cite{kongEventVPREndtoEndWeakly2022} 
backbones using the corresponding FNN model 
from~\cite{lemaireAnalyticalEstimationSpiking2023}. In 
this model, all input activations are dense and every 
synapse requires a full multiply-accumulate (MAC) operation, 
with the associated memory accesses computed from the layer 
dimensions. To account for the sparsity naturally induced 
by ReLU non-linearities in the ANN baselines, the fraction 
of zero-valued activations ($\gamma$) was measured 
empirically for each model by running inference over the 
test set and recording the output sparsity after each ReLU 
layer.

\subsubsection*{Technology assumptions}

All estimations use energy costs for 45\,nm CMOS technology 
at 32-bit precision, drawn from Jouppi 
\textit{et al.}~\cite{jouppiTenLessonsThree2021} as adopted 
in~\cite{lemaireAnalyticalEstimationSpiking2023}: 
$E_{\text{ADD}} = 0.1$\,pJ for a single addition and 
$E_{\text{MUL}} = 3.1$\,pJ for a single multiplication, 
yielding $E_{\text{MAC}} = 3.2$\,pJ and 
$E_{\text{AC}} = 0.1$\,pJ. SRAM access energy is computed 
as a function of memory size via linear interpolation 
(8\,kB\,$\rightarrow$\,10\,pJ, 
32\,kB\,$\rightarrow$\,20\,pJ, 
1\,MB\,$\rightarrow$\,100\,pJ)~\cite{lemaireAnalyticalEstimationSpiking2023}. 
Static power consumption and inter-layer communication 
energy are excluded from this analysis, consistent with 
both reference 
frameworks~\cite{dampfhofferAreSNNsReally2023, 
lemaireAnalyticalEstimationSpiking2023}.

\section*{Acknowledgments}

This work was supported by the French Defense Innovation Agency (AID) under grant number 2023 65 0082. 

\section*{Code availability}
The complete source code for SpikeVPR is available at: \url{https://github.com/GeoffroyK/SpikeVPR}.

\newpage
\bibliographystyle{unsrt}
\bibliography{references}

\newpage

\appendix
\section*{Appendix}


\setcounter{figure}{0}
\renewcommand{\figurename}{Supplementary Figure}

\section{Illustration of the VPR problem}

\begin{figure}[H]
    \centering
    \includegraphics[width=\columnwidth ]{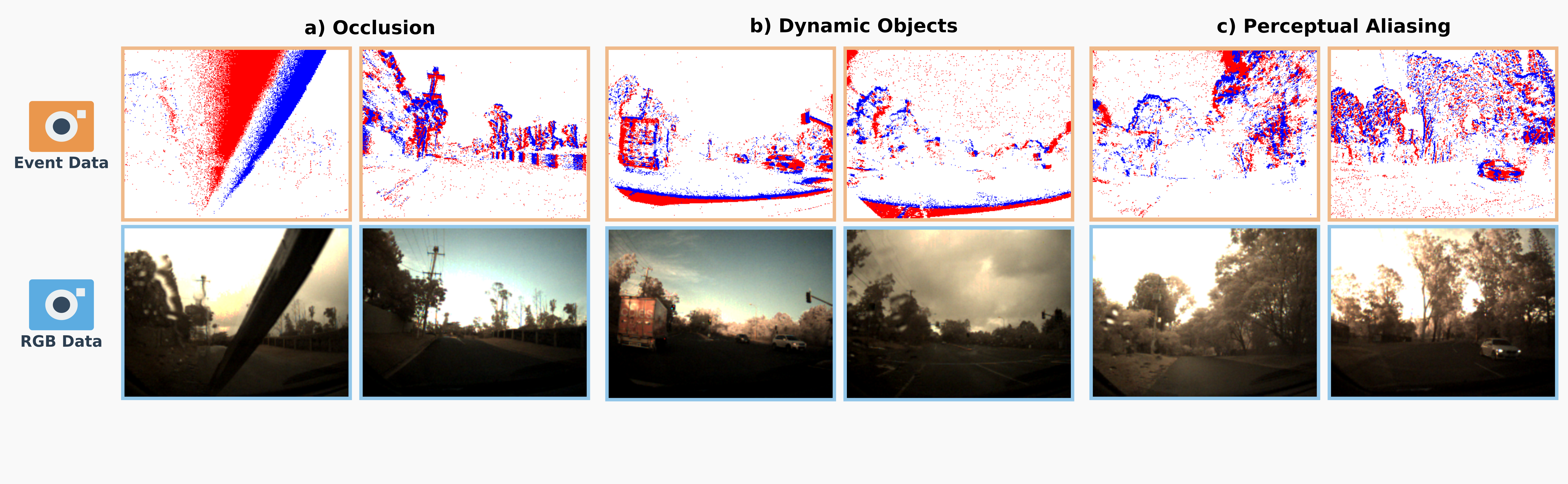}
    \caption{Illustration of three core challenges in VPR, shown for both event-based and RGB modalities on the Brisbane-Event-VPR Dataset.
(a) Occlusion: The same place is captured with and without a significant foreground obstruction (a vehicle windshield pillar), leading to large differences in visual input.
(b) Dynamic Objects: A parked truck partially occludes the scene in one traverse, generating spurious high-activity regions in the event frame that are absent in the reference, even though the underlying place remains the same.
(c) Perceptual Aliasing: Two structurally similar but geographically distinct locations (1.4 km apart) produce similar appearances, creating a risk of false matches for VPR systems. In event data, red and blue spikes indicate brightness increases and decreases, respectively. RGB frames are shown for illustration only and are not used in our study.}
    \label{fig:annex1}
\end{figure}

\begin{figure}[H]
    \centering
    \includegraphics[width=\columnwidth ]{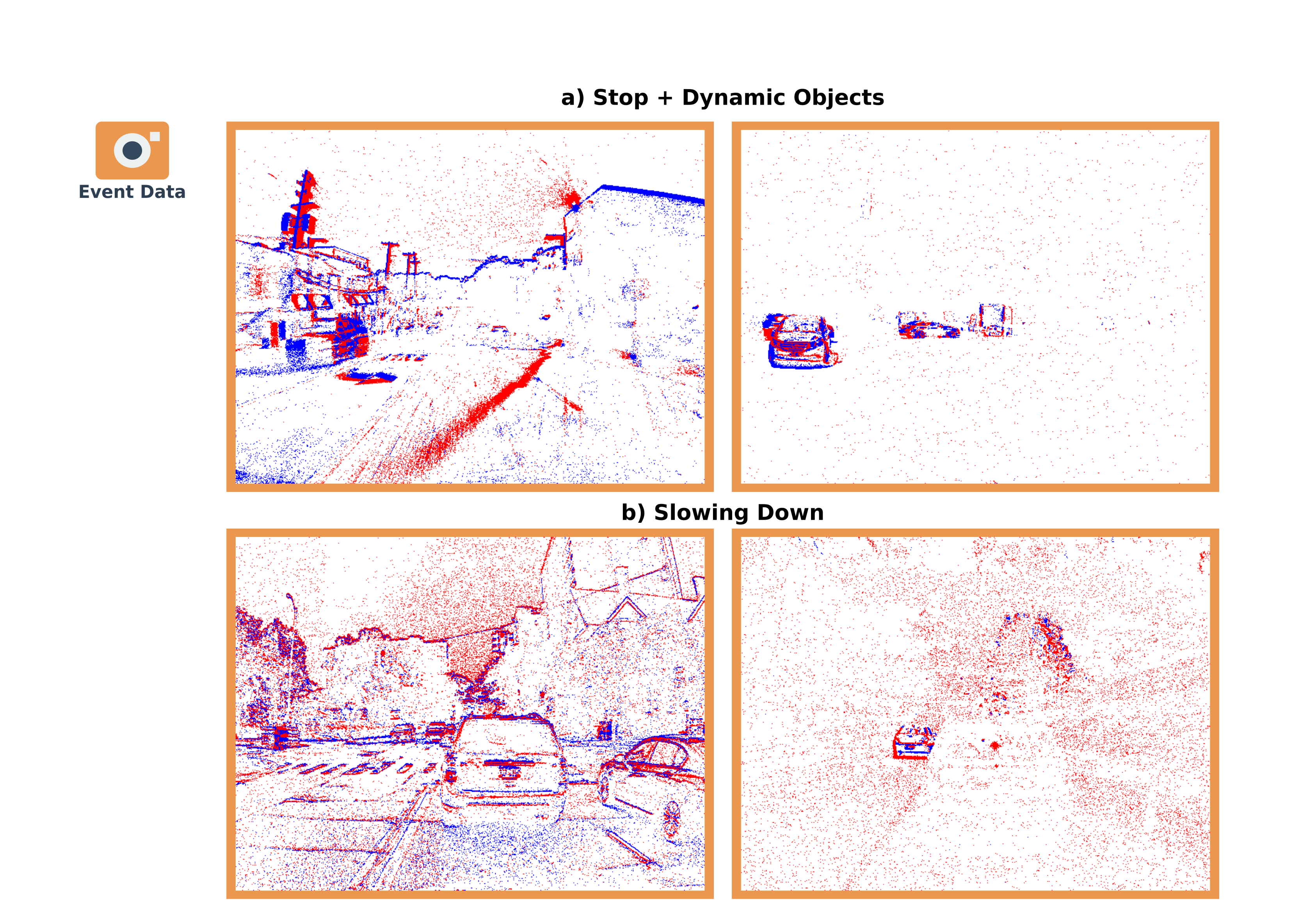}
    \caption{Illustration from the NSAVP dataset showing how the motion of a vehicle equipped with an event-based camera affects VPR. Each pair of images depicts the same location captured during two different traversals.
(a) Stop with moving objects: When the vehicle is stationary, static scene elements generate no events, leaving only moving objects (e.g., passing cars) visible. This causes a near-complete loss of place-descriptive information, with dynamic distractors dominating the event representation.
(b) Slowing Down: As the vehicle decelerates, event rates drop, producing sparse and noisy frames with reduced scene coverage compared to normal-speed traverses.
In both cases, the event representation of the same place differs markedly between traverses, highlighting the sensitivity of event-based descriptors to vehicle velocity changes and dynamic objects.}
    \label{fig:annex2}
\end{figure}

\section{Qualitative results on Brisbane}
\begin{figure}[H]
    \centering
    \includegraphics[width=\textwidth ]{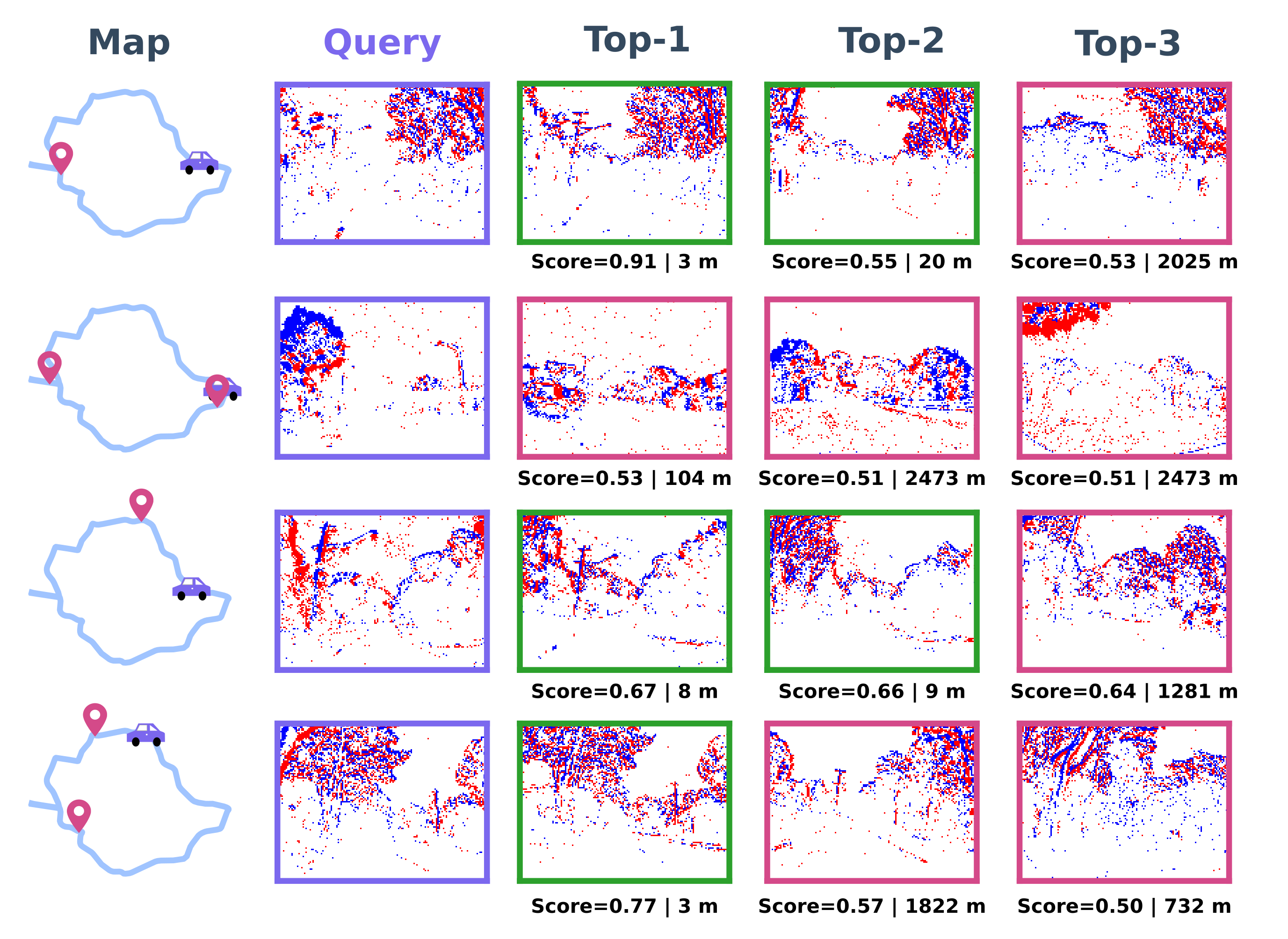}
    \caption{Examples of SpikeVPR predictions on the Brisbane-Event-VPR dataset. For each query (shown in purple on the left), the top three retrieved places are displayed. For each of these places, we report the cosine similarity score and the euclidean distance (in meters) to the query. Correct matches (within 30 meters) are shown in green, whereas incorrect matches are shown in red.}
    \label{fig:annex3}
\end{figure}

\section{SpikeVPR performance with and without pretraining}
\begin{figure}[H]
    \centering
    \includegraphics[width=\textwidth ]{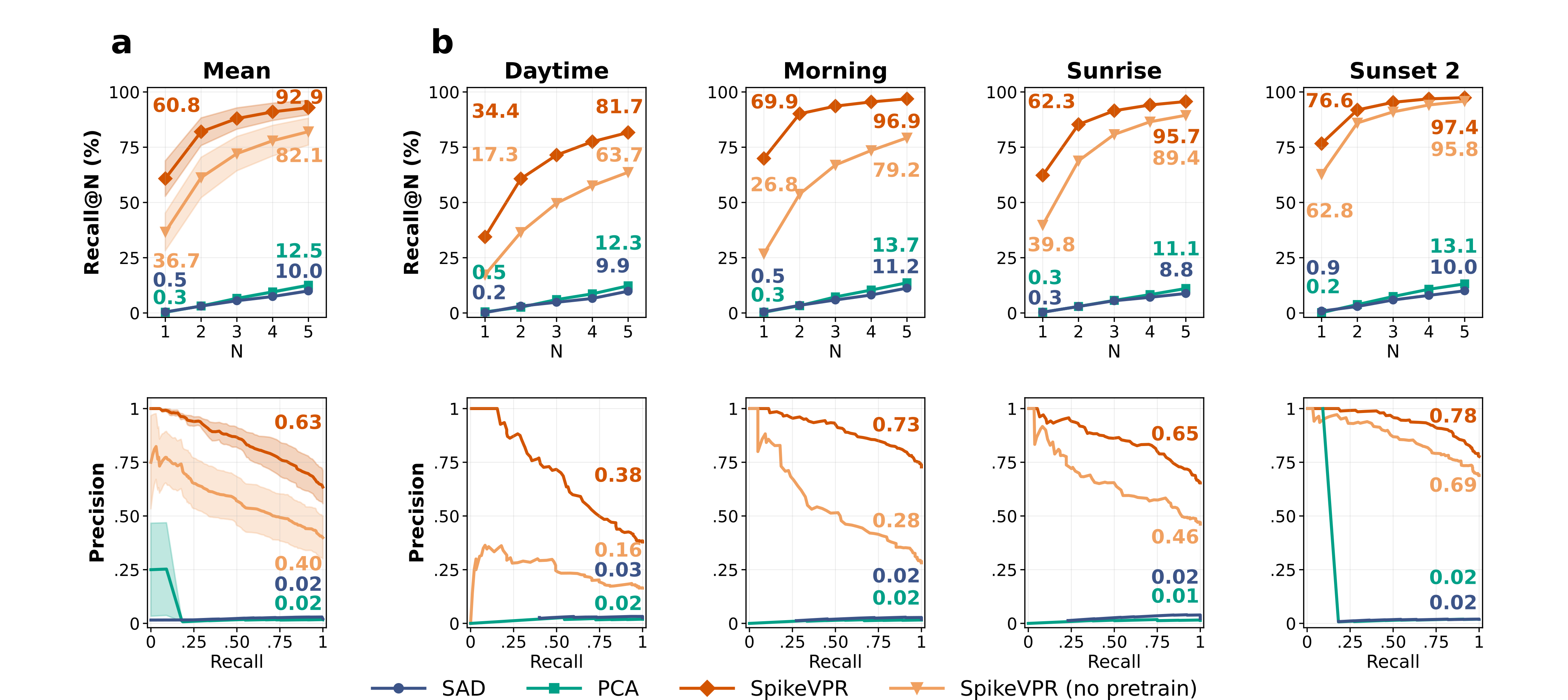}
    \caption{Visual place recognition performance on the Brisbane-Event-VPR dataset with and without pretraining on NSAVP dataset. (\textbf{a}), Mean Recall@N (top) and Precision-Recall curves (bottom) across all four traverses, with shaded regions indicating ± 1 standard error (n = 4 traverses). (\textbf{b}), Per-traverse Recall@N (top) and Precision-Recall curves (bottom) for Daytime, Morning, Sunrise and Sunset conditions. SpikeVPR is evaluated both with pretraining on the NSAVP dataset (dark orange) and without pretraining (light orange), alongside SAD (blue) and PCA (teal) baselines. Values at curve endpoints denote Recall@1, Recall@5 and final precision, respectively.
    }
    \label{fig:annex4}
\end{figure}

\section{Qualitative results on NSAVP}
\begin{figure}[H]
    \centering
    \includegraphics[width=\textwidth ]{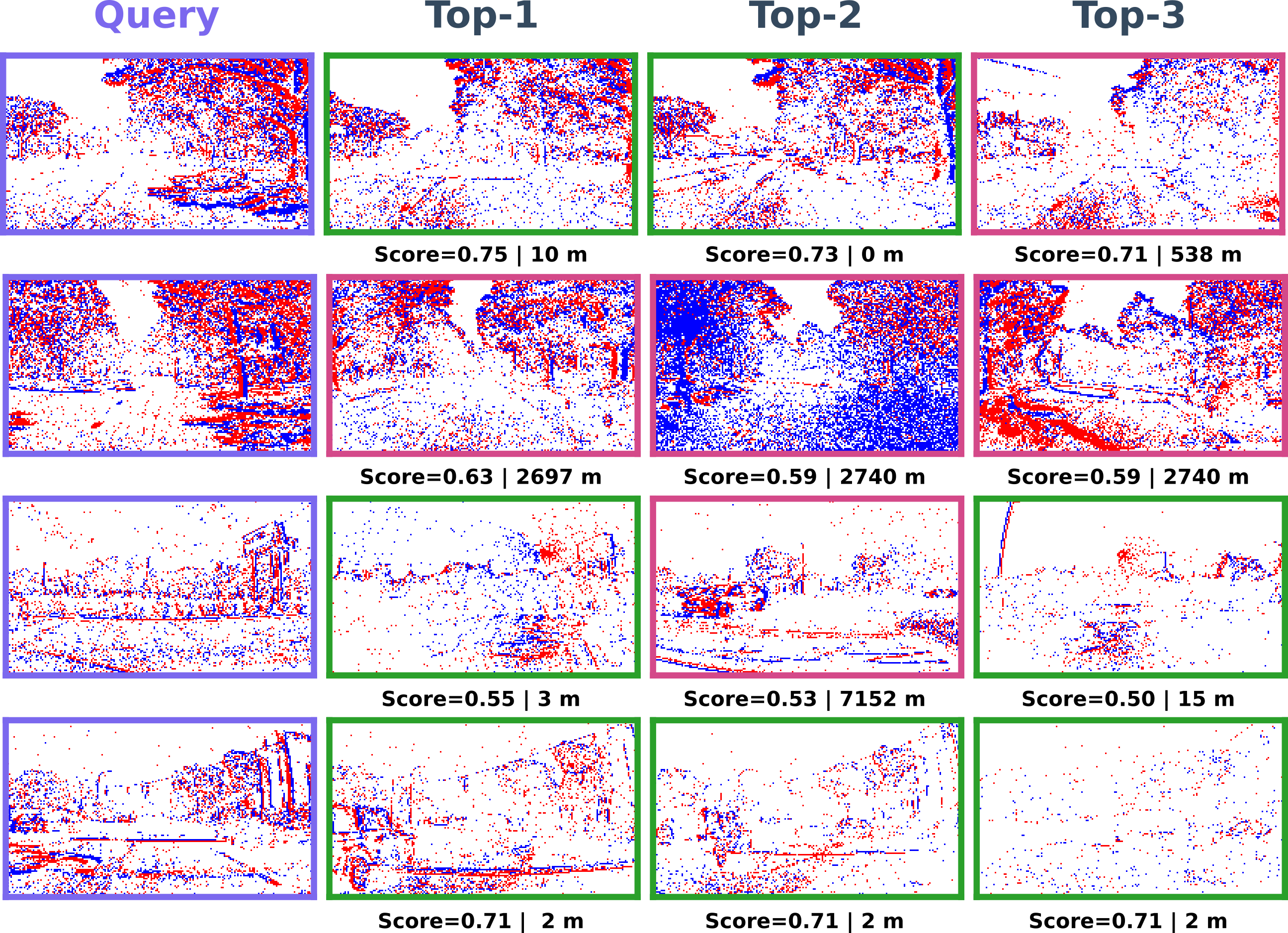}
    \caption{Examples of SpikeVPR predictions on the NSAVP dataset. For each query (shown in purple on the left), the top three retrieved places are displayed. For each of these places, we report the cosine similarity score and the spatial distance (in meters) to the query. Correct matches (within 30 meters) are shown in green, whereas incorrect matches are shown in red.}
    \label{fig:annex5}
\end{figure}

\end{document}